\RequirePackage{fix-cm}
\documentclass[twocolumn,natbib]{svjour3}          
\smartqed  
\usepackage{times}
\usepackage{epsfig}
\usepackage{graphicx}
\usepackage{amsmath}
\usepackage{amssymb}
\usepackage{multirow}
\usepackage{booktabs}
\usepackage{threeparttable}
\usepackage{subfig}
\usepackage{color}
\usepackage{eucal}
\begin{document}

\title{Progressive DARTS: Bridging the Optimization Gap for \textit{NAS in the Wild}
}


\author{Xin Chen         \and
        Lingxi Xie \and
        Jun Wu \and
        Qi Tian
}


\institute{X. Chen and J. Wu \at
              Tongji University \\
              \email\textsf{\{1410452, wujun\}@tongji.edu.cn}           
           \and
           L. Xie and Q. Tian \at
           Huawei Noah's Ark Lab \\
           \email\textsf{198808xc@gmail.com, tian.qi1@huawei.com}
}

\date{Received: date / Accepted: date}

\maketitle

\begin{abstract}
With the rapid development of neural architecture search (NAS), researchers found powerful network architectures for a wide range of vision tasks. However, it remains unclear if the searched architecture can transfer across different types of tasks as manually designed ones did. This paper puts forward this problem, referred to as \textbf{NAS in the wild}, which explores the possibility of finding the optimal architecture in a proxy dataset and then deploying it to mostly unseen scenarios.

We instantiate this setting using a currently popular algorithm named differentiable architecture search (DARTS), which often suffers unsatisfying performance while being transferred across different tasks. We argue that the accuracy drop originates from the formulation that uses a super-network for search but a sub-network for re-training. The different properties of these stages have resulted in a significant \textbf{optimization gap}, and consequently, the architectural parameters ``over-fit" the super-network. To alleviate the gap, we present a progressive method that gradually increases the network depth during the search stage, which leads to the Progressive DARTS (P-DARTS) algorithm. With a reduced search cost ($7$ hours on a single GPU), P-DARTS achieves improved performance on both the proxy dataset (CIFAR10) and a few target problems (ImageNet classification, COCO detection and three ReID benchmarks). Our code is available at \textsf{https://github.com/chenxin061/pdarts}.


\end{abstract}

\section{Introduction}\label{intro}

Recently, the research progress of computer vision has been largely boosted by deep learning~\citep{lecun2015deep}. The core part of deep learning is to design and optimize deep neural networks, for which a few popular models have been manually designed and achieved state-of-the-art performance at that time~\citep{krizhevsky2012imagenet, szegedy2015going, he2016deep, zhang2018shufflenet, howard2017mobilenets}. However, designing neural network architectures requires both expertise and heavy computational resources. The appearance of neural architecture search (NAS) has changed this situation, which aims to discover powerful network architectures automatically and has achieved remarkable success in image recognition~\citep{zoph2016neural,zoph2018learning,liu2018progressive,tan2019efficientnet}.


\begin{figure*}[t]
\centering
\subfloat[NAS in the Wild]{
    \begin{minipage}[b]{0.63\linewidth}
    \includegraphics[width=0.99\linewidth]{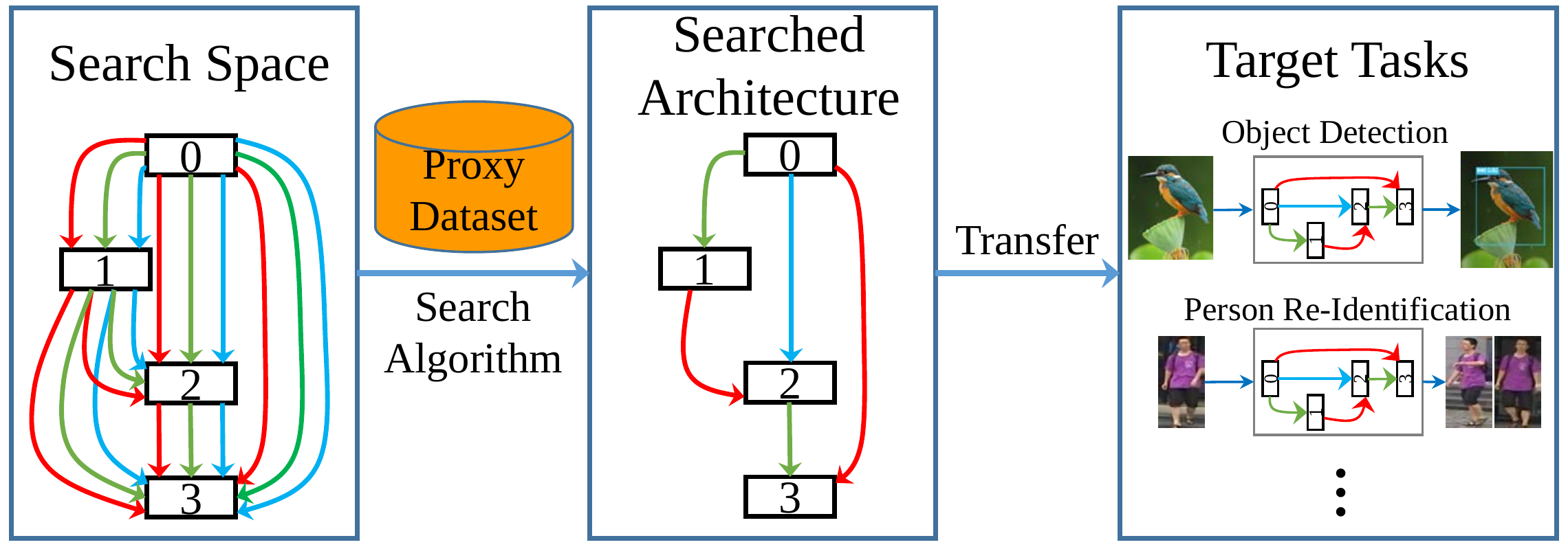}
    \end{minipage}}
\subfloat[Our Solution: P-DARTS]{
    \begin{minipage}[b]{0.36\linewidth}
    \begin{center}
    \includegraphics[width=0.9\linewidth]{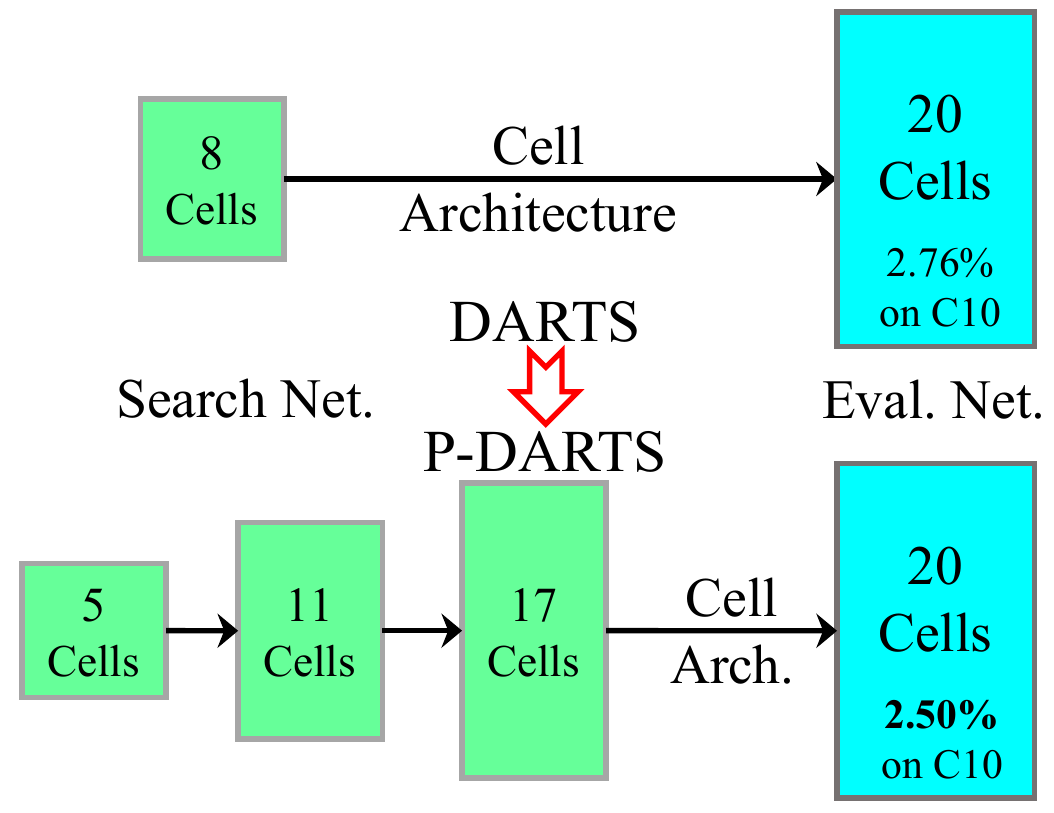}
    \end{center}
    \end{minipage}}
\caption{Left: the setting of \textbf{NAS in the wild}, which aims to transfer the optimal architecture found in the proxy dataset to unknown scenarios. Right: our solution, P-DARTS (bottom), bridges the \textbf{optimization gap} between architecture search and evaluation by gradually increasing the depth of the super-network. 
DARTS (top) is also placed here for comparison. Green and blue indicate search and evaluation, respectively.}
\label{motivation}
\end{figure*}

In the early age of NAS, researchers focused on heuristic search methods, which sample architectures from a large search space and perform individual evaluations. Such approaches, while being safe in finding powerful architectures, require massive computational overheads~\citep{zoph2016neural,real2018regularized, zoph2018learning}. To alleviate this burden, researchers have designed efficient approaches to reuse computation in the searched architectures~\citep{cai2018efficient}, which was later developed into constructing a super-network to cover the entire search space~\citep{pham2018efficient}. Among them, DARTS~\citep{liu2018darts} is an elegant solution that relaxes the discrete search space into a continuous, differentiable function. Thus, the search process requires optimizing the super-network and can be finished within GPU-hours.

Despite the efficiency of super-network-based search methods, most of them suffer from the issue of instability, which indicates that (i) the accuracy can be sensitive to random initialization, and (ii) the searched architecture sometimes incurs unsatisfying performance in other datasets or tasks. While directly searching over the target problem is always a solution, we argue that studying this topic may unleash the potentials of NAS. To this end, we formalize a setting named \textbf{NAS in the wild}, illustrated in Figure~\ref{motivation}, which advocates for the searched architecture on any proxy dataset to be easily deployed to different application scenarios.

We argue that the instability issue originates from that the search stage fits the \textit{super-network} on the proxy dataset, but the re-training stage actually applies the optimal \textit{sub-network} to either the same dataset or a different task. Even if the proxy dataset and the target dataset are the same, one cannot expect the best super-network, after being pruned, produces the best sub-network. This is called the \textbf{optimization gap}. In this work, we explore a practical method to alleviate the gap, which involves gradually adjusting the super-network so that its properties converge to the sub-network by the end of the search process.

Our approach, named Progressive DARTS (P-DARTS), is built on DARTS, a recently published method for differentiable NAS. As shown in Figure~\ref{motivation}(b), the search process of P-DARTS is divided into multiple stages, and the depth of the super-network gets increased at the end of each stage. This brings two technical issues, and we provide solutions accordingly. First, since heavier computational overheads are required when searching with a deeper super-network, we propose \textbf{search space approximation}, which reduces the number of candidates (operations) when the network depth is increased. Second, optimizing a deep super-network may cause unstable gradients, and thus the search algorithm is biased heavily towards \textit{skip-connect}, a learning-free operator that often falls on a rapid direction of gradient decay. Consequently, it reduces the learning ability of the found architecture, for which we propose \textit{search space regularization}, which (i) introduces operation-level Dropout~\citep{srivastava2014dropout} to alleviate the dominance of \textit{skip-connect} during training, and (ii) regularizes the appearance of \textit{skip-connect} when determining the final sub-network.

The effectiveness of P-DARTS is firstly verified on the standard vision setting, \textit{i.e.}, searching and evaluating the architecture on the CIFAR10 dataset. We achieve state-of-the-art performance (a test error of $2.50\%$) on CIFAR10 with $3.4$M parameters. In addition, we demonstrate the benefits of search space approximation and regularization: the former reduces the search cost to $0.3$ GPU-days on CIFAR10, surpassing ENAS~\citep{pham2018efficient}, an approach known for search efficiency; the latter largely reduces the fluctuation of individual search trials and thus improving its reliability. Next, we investigate the application in the wild, in which the searched architecture on CIFAR10 transfers well to CIFAR100 classification, ImageNet classification, COCO detection, and three person re-identification (ReID) tasks, \textit{e.g.}, on ImageNet, it achieves top-$1$/$5$ errors of $24.4\%$/$7.4\%$, respectively, comparable to the state-of-the-art under the mobile setting. Furthermore, architecture search is also performed on ImageNet, and the discovered architecture shows superior performance. 

The preliminary version of this work appeared as~\citep{chen2019progressive}. In this journal version, we extend the original work by several aspects. First, we present the new setting of \textit{NAS in the wild}, which provides a benchmark for evaluating the generalization ability of NAS approaches. Second, we complement a few diagnostic experiments to further reveal that bridging the optimization gap is helpful to accomplish the goal of NAS in the wild. Third, we extend the search method so that it can directly search on ImageNet and thus produce more powerful architectures for large-scale image recognition.

The remaining part of this paper is organized as follows. Section~\ref{rw} briefly introduces related work to our research. Then, Section~\ref{problem} illustrates the problem, NAS in the wild, and Section~\ref{method} elaborates the optimization gap and the P-DARTS approach. After extensive experiments are shown in Section~\ref{exp}, we conclude this work in Section~\ref{conclusion}.

\section{Related Work}\label{rw}

Image recognition is a fundamental task in computer vision.
In recent years, with the development of deep learning, CNNs have been dominating image recognition~\citep{krizhevsky2012imagenet, simonyan2014very, he2016deep}. 
A few elaborately designed handcrafted architectures have been proposed, including VGGNet~\citep{simonyan2014very}, ResNet~\citep{he2016deep}, DenseNet~\citep{huang2017densely}, \textit{etc.}, all of which highlighted the importance of human experts in network design. 

In the era of hand-designed architectures, the main roadmap of architecture design resided in how to enlarge the depth of CNNs efficiently. AlexNet~\citep{krizhevsky2012imagenet} proposed to use the ReLU activation function and Local Response Normalization (LRN) to alleviate the gradient diffusion and achieved the state-of-the-art performance on ImageNet classification at that time. VGGNet~\citep{simonyan2014very} proposed to stack convolutions with identical small kernel size and initialize deeper networks with previously learned weights of a shallow work, which resulted in a network of $19$ layers. GoogLenet~\citep{szegedy2015going} introduced to connect convolutions with different kernel sizes in parallel, which led to a reduction of network parameters, an increase of network depth, and a promotion on parameter utilization.  In ResNet~\citep{he2016deep}, the depth of networks was further increased to $152$ layers for ImageNet and even $1\rm{,}202$ layers for CIFAR10, with the help of the newly proposed skip connection and residual block. 
After that, DenseNet~\citep{huang2017densely} inserted skip connection between all layers in the building block to formulate a densely connected CNN, which largely strengthened information propagation and feature reutilization. 
Apart from this depth route, network width was also a critical aspect of performance promotion. WRN~\citep{zagoruyko2016wide} explored the possibility of scaling up the network width of ResNet and achieved brilliant results. PyramidNet~\citep{han2017deep} extended this idea to design a pyramid-like ResNet, which further promoted the network capability. 

This work is in the category of the emerging field of neural architecture search, a process of automating architecture engineering technique~\citep{elsken2018neural}. 
In the early 2000s, pioneer researchers attempted to generate better topology automatically with evolutionary algorithms~\citep{stanley2002evolving}.
Early NAS works tried to search for basic components and topology of neural networks to construct a complete network~\citep{baker2016designing, suganuma2017genetic, xie2017genetic}, while recent works focused on finding robust cells~\citep{zoph2018learning, real2018regularized, dong2019searching}. 
Among these works, heuristic algorithms were widely adopted in the NAS pipeline. 
Baker {\em et al.}~\citep{baker2016designing} firstly applied reinforcement learning (RL) to neural architecture search and adopted an RNN-based controller to guide the sampling process for the network configuration.  
\citep{xie2017genetic} encoded the architecture of a CNN into binary codes and used a general evolutionary algorithm to evolve for a better global network topology. 
Considering the weakness of the scalability of a global network architecture, \citep{zoph2016neural} adopted RL to search for the configuration of building blocks, which are also referred to as cells. \citep{real2018regularized} proposed to regularize the standard evolutionary algorithm in the NAS pipeline with aging evolution and, for the first time, surpassed the best manually designed architectures on image recognition. 

A critical drawback of the above approaches is the expensive search cost ($3\rm{,}150$ GPU-days for EA-based AmoebaNet~\citep{real2018regularized} and $20\rm{,}000$ GPU-days for RL-based NASNet~\citep{zoph2016neural}), because their methods require to sample and evaluate numerous architectures by training them from scratch.
There were two lines of solutions.
The first one involved reducing the search space~\citep{zoph2018learning}, and the second one optimized the exploration policy (\textit{e.g.}, learning a surrogate model~\citep{liu2018progressive}) in the search space so that the search process becomes more efficient.

Recently, search efficiency has become one of the main concerns on NAS, and the search cost was reduced to a few GPU-days with the help of weight sharing technique~\citep{pham2018efficient, liu2018darts}. 
In this pipeline, a super-network that contains all candidate architectures in the search space is trained, and sub-architectures are evaluated with shared weights from the super-network. 
ENAS~\citep{pham2018efficient} proposed to adopt a parameter sharing scheme among child models to bypass the time-consuming process of candidate architecture evaluation by training them from scratch, which dramatically reduced the search cost to less than one GPU-day. 
DARTS~\citep{liu2018darts} introduced a differentiable NAS framework to relax the discrete search space into a continuous one by weighting candidate operations with architectural parameters, which achieved comparable performance and remarkable efficiency improvement compared to previous approaches. 
Following DARTS, GDAS~\citep{dong2019searching} proposed to use the Gumbel-softmax sampling trick to guide the sub-graph selection process. 
With the BinaryConnect scheme, ProxylessNAS~\citep{cai2018proxylessnas} adopted the differentiable framework and proposed to search architectures on the target task instead of adopting the conventional proxy-based framework. 
A main drawback of DARTS-based approaches is the instability issue caused by the optimization gap depicted in Section~\ref{intro}. 
SNAS~\citep{xie2018snas} proposed to constrain the architecture parameters to be one-hot to tackle the inconsistency in optimizing objectives between search and evaluation scenarios, which can be regarded as an attempt of reducing the optimization gap. However, SNAS reported only comparable classification performance to DARTS on both proxy and target datasets.

\section{Problem: \textit{NAS in the Wild}}\label{problem}

We investigate the setting of \textit{NAS in the Wild}, which seeks for a NAS algorithm that can search in a \textit{proxy} dataset and freely transfer to a wide range of target datasets or even other types of recognition tasks. This is important for real-world scenarios, as there may not be sufficient resources, in terms of either data or computation, for a complete NAS process to be executed.

Note that the community has witnessed a few recent works, sometimes referred to as \textit{proxyless} NAS~\citep{cai2018proxylessnas}, in searching neural architectures on the target dataset directly. Our setting does not contradict these efforts, and we argue that both settings have their own advantages. On the one hand, searching on the target dataset directly enables more accurate properties of the specified dataset to be captured and, most often, leads to improved performance on the target dataset. On the other hand, we desire the ability of directly transferring the searched architecture to other scenarios. This task not only makes it easier in application, and also raises new challenges which we believe beneficial for the research field of NAS.

The most significant difficulty brought by this setting is the enlarged gap between the search stage and the evaluation stage, which we will elaborate in detail in Section~\ref{method:gap}. In this paper, we present a practical solution that largely shrinks this gap and thus improves the ability of model transfer.

\section{Method: Progressive DARTS}\label{method}

\begin{figure*}[t]
\begin{center}
   \includegraphics[width=0.95\linewidth]{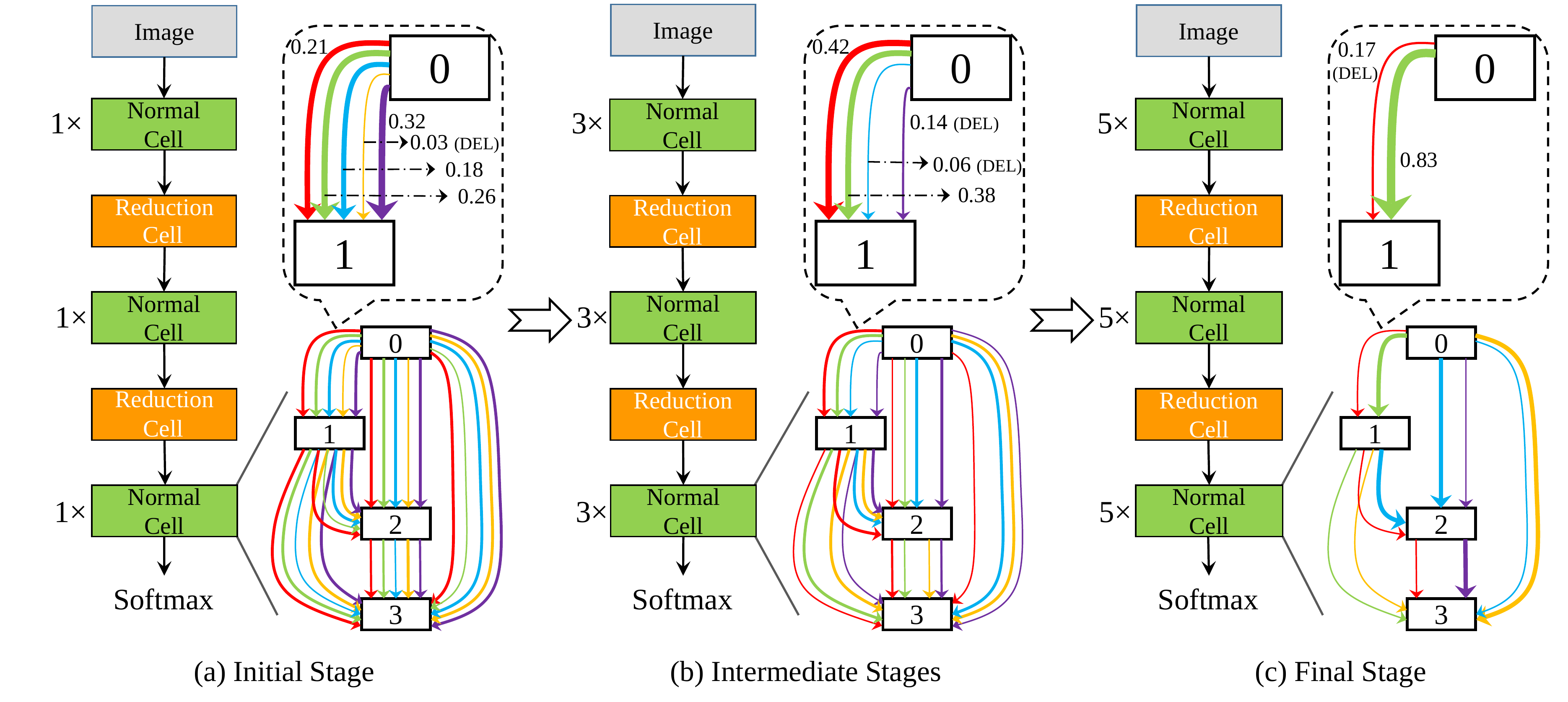}
\end{center}
\caption{The overall pipeline of P-DARTS (best viewed in color). For simplicity, only one intermediate stage is shown, and only the normal cells are displayed. The depth of the super-network increases from $5$ at the initial stage to $11$ and $17$ at the intermediate and final stages, while the number of candidate operations (shown in connections with different colors) is shrunk from $5$ to $4$ and $2$ accordingly. The lowest-scored ones at the previous stage are dropped (the scores are shown next to each connection). We obtain the final architecture by considering the final scores and possibly additional rules.}
\label{pipeline}
\end{figure*}

\subsection{Preliminary: Differentiable Architecture Search}

Our work is based on DARTS~\citep{liu2018darts}, which adopts a cell-based search framework that searches for robust architectures building blocks, {\em i.e.}, cells, and then stacks searched cells orderly for $L$ times to construct the target network. 
Thus, the search space is represented in the form of cells. A cell is denoted as a directed acyclic graph (DAG) $\mathcal{G}$ and composed of $N$ nodes (vertexes) and their corresponding edges. A node $x_i$ represents a feature layer, {\em i.e.}, the output of a specific operation. 
The first two nodes of a cell are the input nodes, which come from the outputs of previous cells or stem convolutions located at the beginning of the network. 
We denote the operation space as $\mathcal{O}$, in which each element represents a candidate operation (mathematical function) $o(\cdot)$. 
An intermediate node $x_j$ is connected to all of its preceding nodes $\{x_1, x_2, ..., x_{j-1}\}$ with edge $\mathrm{E}_{(i, j)} (i < j)$, where operations from the operation space are used to link the information flow between node $x_i$ and $x_j$. 
To relax the discrete search space to be continuous, operations on each edge are weighted with a set of architectural parameters $\alpha^{(i, j)}$, which is normalized with the Softmax function and is thus formulated as: 
\begin{equation}\label{eq1}
\centering
  f_{i, j}(x_i) = \sum_{o\in\mathcal{O}_{i, j}}{\frac{\mathrm{exp}(\alpha_o^{(i,j)})}{\sum_{o'\in\mathcal{O}}\mathrm{exp}(\alpha_{o'}^{(i,j)})}o(x_i)}.
\end{equation}
All feature maps passed into the intermediate node $x_j$ are integrated together by summation, denoted as  $x_j = \sum_{i < j}{f_{i,j}(x_i)}$. 
The output node is defined as $x_{N-1} = \mathrm{concat}(x_2, x_3, \cdots, x_{N-2})$, where $\mathrm{concat}(\cdot)$ concatenates all input signals in the channel dimension. 

The architectural parameters in DARTS are jointly optimized with the network parameters, \textit{i.e.}, the convolutional weights. The output architecture is generated by operation pruning according to the learned architectural parameters, with at most one non-\textit{zero} operation on a specific edge and two preserved edge for each intermediate node. 
For more technical details, please refer to the original DARTS paper~\citep{liu2018darts}. 

\subsection{The Optimization Gap}\label{method:gap}

The most significant drawback of DARTS, especially when discussed in the scenario of \textit{NAS in the wild}, lies in the gap between search and evaluation. To be specific, in DARTS as well as other super-network-based NAS approaches, the search goal is to optimize the objective function with respect to network parameters, $\omega$, and the architectural parameters, $\alpha$, on the proxy dataset, $\mathcal{D}_\mathrm{S}$. The overall objective can be written as:
\begin{equation}\label{eq2}
\centering
\begin{aligned}
    \alpha^\star={\arg\min_{\alpha}}\left\{{\min_{\omega_\mathrm{S}}}\mathcal{L}_\mathrm{S}(\omega_\mathrm{S},\alpha;\mathcal{D}_\mathrm{S}, \mathcal{C}_\mathrm{S})\right\},
\end{aligned}
\end{equation}
where $\mathcal{C}_\mathrm{S}$ denotes the network configuration for architecture search. 
The output of the search process is an optimal sub-network $\mathcal{A}$ generated according to the learned, optimal architectural parameter, $\alpha^\star$.
At the evaluation phase, the destination is to train the evaluation network constructed with the searched architecture $\mathcal{A}$ on the target dataset $\mathcal{D}_\mathrm{E}$ for a better performance, thus we have:
\begin{equation}\label{eq2}
\centering
\begin{aligned}
    \omega_\mathrm{E}^\star=\arg\min_{\omega_\mathrm{E}}\mathcal{L}_\mathrm{E}(\omega_\mathrm{E};\mathcal{A}, \mathcal{D}_\mathrm{E}, \mathcal{C}_\mathrm{E}),
\end{aligned}
\end{equation}
where $\omega_\mathrm{E}$ and $\mathcal{C}_\mathrm{E}$ denote the network parameters and configuration at evaluation time, respectively.

There are severe mismatch problems that happen to the DARTS framework between search and evaluation scenarios (mismatch between $\mathcal{C}_\mathrm{S}$ and $\mathcal{C}_\mathrm{E}$) in network shape, hyper-parameters, training policies, \textit{etc.}. We summarize these problems into the \textbf{optimization gap} between training the super-network and applying the sub-network to the target network. For example, a typical optimization gap comes from the inconsistency of the operation pruning process, since the objective of the super-network is to jointly optimize network weights $\omega_\mathrm{S}$ of all candidate operations and the architectural parameters $\alpha$, while the objective of training the target network is only to optimize the network weights $\omega_\mathrm{E}$ of a few selected operations. These mismatches result in dramatic performance deterioration when the discovered architectures are applied to real-world applications. In particular, the difference between the proxy and target dataset and/or task can even enlarge the optimization gap, and thus cause difficulties of applying the searched architecture in the wild.

\subsection{Progressive Search to Bridge the Depth Gap}

Among the optimization gaps, that caused by different network depths is one of the main sources of performance deterioration. We propose to alleviate it by progressively increasing the search depth, which is built upon our search space approximation scheme. 
Besides, the mismatch on network width, \textit{i.e.}, the number of channels of feature maps, is also an essential factor associated with performance when searching architectures on large and complex datasets, and we tackle it by progressively increasing search width. 

To be specific, the architecture search process of DARTS is performed on a super-network with $8$ cells, and the discovered architecture is evaluated on a network with either $20$ cells (on CIFAR10) or $14$ cells (on ImageNet).
There is a considerable difference between the behavior of shallow and deep networks~\citep{ioffe2015batch, srivastava2015training, he2016deep}, which implies that the architectures we discovered in the search process are not necessarily the optimal one for evaluation.
We name this the \textbf{depth gap} between search and evaluation.
To verify it, we executed the search process of DARTS for multiple times and found that the normal cells of discovered architectures tend to keep shallow connections instead of deep ones, {\em i.e.}, the search algorithm prefers to preserve those edges connected to the input nodes instead of cascading between intermediate nodes.
This is because shallow networks often enjoy faster gradient descent during the search process. However, such property contradicts the common sense that deeper networks tend to perform better~\citep{simonyan2014very,szegedy2015going,he2016deep,huang2017densely}. Therefore, we propose to bridge the depth gap with the strategy that progressively increases the network depth during the search process,
so that at the end of the search, the depth of the super-network is sufficiently close to the network configuration used in the evaluation.
Here we adopt a progressive manner, instead of directly increasing the search depth to the target level, because we expect to search in shallow networks to reduce the search space with respect to candidate operations, so as to alleviate the risk of search in deep networks.
The effectiveness of this progressive strategy will be verified in Section~\ref{comp_depth}.

The difficulty comes from two aspects.
First, the computational overhead increases linearly with the search depth, which brings issues in both time expenses and computational overheads.
In particular, in DARTS, GPU memory usage is proportional to the depth of the super-network.
The limited GPU memory forms a major obstacle, and the most straightforward solution is to trim the channels number in each operation -- DARTS~\citep{liu2018darts} tried it but reported a slight performance deterioration, because it enlarged the mismatch on network width, another aspect of the optimization gap. 
To address this problem, we propose a \textbf{search space approximation} scheme to progressively reduce the number of candidate operations at the end of each stage according to the architectural parameters, the scores of operations in the previous stage as the criterion of selection.
Details of search space approximation are presented in Section~\ref{ssa}.

Second, we find that when searching on a deeper super-network, the differentiable approaches tend to bias towards the \textit{skip-connect} operation,
because it accelerates forward and backward propagation and often leads to the fastest route of gradient descent. 
However, this operation is parameter-free, which implies a relatively weak ability to learn visual representations. 
To this end, we propose another scheme named \textbf{search space regularization}, which adds an operation-level Dropout~\citep{srivastava2014dropout} to prevent the architecture from ``over-fitting" and restricts the number of preserved \textit{skip-connects} for further stability.
Details of search space regularization are presented in Section~\ref{ssr}.

\subsubsection{Search Space Approximation}
\label{ssa}

A toy example is presented in Figure~\ref{pipeline} to demonstrate the idea of search space approximation. 
The entire search process is split into multiple stages, including an initial stage, one or a few intermediate stages, and a final stage. 
For each stage, $\mathfrak{S}_k$, the super-network is constructed with $L_k$ cells and the operation space consists of $O_k$ candidate operations, \textit{i.e.},  $|\mathcal{O}_{(i, j)}^k| = O_k$.

According to our motivation, the super-network of the initial stage is relatively shallow, but the operation space is large ($\mathcal{O}_{(i, j)}^1 \equiv \mathcal{O}$).
After each stage, $\mathfrak{S}_{k-1}$, the architectural parameters $\alpha_{k-1}$ are optimized and the scores of the candidate operations on each edge are ranked according to the learned $\alpha_{k-1}$. 
We increase the depth of the super-network by stacking more cells, \textit{i.e.}, $L_{k}>L_{k-1}$, and approximate the operation space according to the ranking scores in the meantime.
As a consequence, the new operation set on each edge $\mathcal{O}_{(i, j)}^k$ has a smaller size than $\mathcal{O}_{(i, j)}^{k-1}$, or equivalently, $O_{k}<O_{k-1}$.
The criterion of approximation is to drop a part of less important operations on each edge, which are specified to be those assigned with a lower weight during the previous stage, $\mathfrak{S}_{k-1}$.
As shown in Table~\ref{mem_usage}, this strategy is memory-efficient, which enables the deployment of our approach on regular GPUs, \textit{e.g.}, with a memory of $16$GB.

The growth of architecture depth continues until it is sufficiently close to that used in the evaluation. 
After the last search stage, the final cell topology (bold lines in Figure~\ref{pipeline}(c)) is derived according to the learned architecture parameters $\alpha_K$.
Following DARTS, for each intermediate node, we keep two individual edges whose largest non-\textit{zero} weights are top-ranked and preserve the most important operation on each retained edge. 

\subsubsection{Search Space Regularization}
\label{ssr}

At the start of each stage, $\mathfrak{S}_k$, we train the (modified) architecture from scratch, \textit{i.e.}, all network weights and architectural parameters are re-initialized randomly, because several candidates have been abandoned on each edge\footnote{We also tried to start with architectural parameters learned from the previous stage, $\mathfrak{S}_{k-1}$, and adjust them according to Eq. \ref{eq1} to ensure that the weights of preserved operations should still sum to one. This strategy reported slightly lower accuracy. Actually, we find that only an average of $5.3$ (out of $14$ normal edges) most significant operations in $\mathfrak{S}_1$ continue to have the largest weight in $\mathfrak{S}_2$, and the number is only slightly increased to $6.7$ from $\mathfrak{S}_2$ to $\mathfrak{S}_3$ -- this is to say, deeper architectures may have altered preferences.}.
However, training a deeper network is harder than training a shallow one~\citep{srivastava2015training}.
In our particular setting, we observe that information prefers to flow through \textit{skip-connect} instead of \textit{convolution} or \textit{pooling}, which is arguably due to the fact that \textit{skip-connect} often leads to rapid gradient descent, especially on small proxy datasets (CIFAR10 or CIFAR100) which are relatively easy to fit. 
The gradient of a \textit{skip-connect} operation with respect to the input is always $1.0$, while that of \textit{convolutions} is much smaller ($\left[10^{-3},10^{-2}\right]$ according to our statistics). Another important reason is that, during the start of training, weights in \textit{convolutions} are less meaningful, which results in unstable outputs compared to \textit{skip-connect} which is weight-free, and such outputs are not likely to have high weights in classification. Both reasons make \textit{skip-connect} accumulate weights much more rapidly than other operations. 
Consequently, the search process tends to generate architectures with many \textit{skip-connect} operations, which limits the number of learnable parameters and thus produces an unsatisfying performance at the evaluation stage.
This is essentially a kind of over-fitting.

We address this problem by search space regularization, which consists of two parts. First, we insert operation-level Dropout~\citep{srivastava2014dropout} after each \textit{skip-connect} operation to partially ``cut off" the straightforward path through \textit{skip-connect}, and facilitate the algorithm to explore other operations.
However, if we constantly block the path through \textit{skip-connect}, the algorithm will unfairly drop them by assigning lower weights to them, which is harmful to the final performance. 
To address this contradiction, we gradually decay the Dropout rate during the training process in each search stage Thus the straightforward path through \textit{skip-connect} is blocked at the beginning and treated equally afterward when parameters of other operations are well learned, leaving the algorithm itself to make the decision. 

Despite the use of operation-level Dropout, we still observe that \textit{skip-connect}, as a special kind of operation, has a significant impact on recognition accuracy at the evaluation stage. 
Empirically, we perform $3$ individual search processes on CIFAR10 with identical search setting, but find that the number of preserved \textit{skip-connects} in the normal cell, after the final stage, varies from $2$ to $4$.
In the meantime, the recognition performance at the evaluation stage is also highly correlated to this number, as we observed before.
This motivates us to design the second regularization rule, architecture refinement, which simply restricts the number of preserved \textit{skip-connect} operations of the final architecture to be a constant $M$.
This is done with an iterative process, which starts with constructing a cell topology using the standard rule described by DARTS.
If the number of \textit{skip-connects} is not exactly $M$, we search for the $M$ \textit{skip-connect} operations with the largest architecture weights in this cell topology and set the weights of others to $0$, then redo cell construction with modified architectural parameters. 

We emphasize that the second regularization technique must be applied on top of the first one, otherwise, in the situations without operation-level Dropout, the search process is producing low-quality architectural weights, based on which we could not build up a powerful architecture even with a fixed number of \textit{skip-connects}.

\begin{table*}[t]
\caption{Comparison with state-of-the-art architectures on CIFAR10 and CIFAR100.
$^\dagger$ indicates that this result is obtained by transferring the corresponding architecture to CIFAR100. $^\ddagger$ We ran the publicly available code with necessary modifications to fit PyTorch 1.0, and a single run took about $0.5$ GPU-days for the first order and $2$ GPU-days for the second order, respectively.}
\begin{center}
\begin{tabular}{lccccc}
\hline\noalign{\smallskip}
\textbf{\multirow{2}{*}{Architecture}} & \multicolumn{2}{c}{\textbf{Test Err. (\%})} & \textbf{Params} & \textbf{Search Cost} & \textbf{\multirow{2}{*}{Search Method}} \\
\cmidrule(lr){2-3}
&                            \textbf{C10} & \textbf{C100} & \textbf{(M)} & \textbf{(GPU-days)} &\\
\noalign{\smallskip}\hline\noalign{\smallskip}
DenseNet-BC~\citep{huang2017densely}                       & 3.46 & 17.18 & 25.6 & -    & manual \\
\noalign{\smallskip}\hline\noalign{\smallskip}
NASNet-A + cutout~\citep{zoph2018learning}                 & 2.65 & -     & 3.3  & 1\rm{,}800 & RL      \\
AmoebaNet-A + cutout~\citep{real2018regularized}           & 3.34 & -     & 3.2  & 3\rm{,}150 & evolution \\
AmoebaNet-B + cutout~\citep{real2018regularized}           & 2.55 & -     & 2.8  & 3\rm{,}150 & evolution \\
Hireachical Evolution~\citep{liu2017hierarchical}          & 3.75 & -     & 15.7 & 300  & evolution \\
PNAS~\citep{liu2018progressive}                            & 3.41 & -     & 3.2  & 225  & SMBO \\
ENAS + cutout~\citep{pham2018efficient}                    & 2.89 & -     & 4.6  & 0.5  & RL \\
\noalign{\smallskip}\hline\noalign{\smallskip}
DARTS (first order) + cutout~\citep{liu2018darts}          & 3.00 & 17.76$^\dagger$ & 3.3 & 1.5$^\ddagger$  & gradient-based \\
DARTS (second order) + cutout~\citep{liu2018darts}         & 2.76 & 17.54$^\dagger$ & 3.3 & 4.0$^\ddagger$ & gradient-based \\
SNAS + mild constraint + cutout~\citep{xie2018snas}        & 2.98 & -     & 2.9  & 1.5  & gradient-based \\
SNAS + moderate constraint + cutout~\citep{xie2018snas}    & 2.85 & -     & 2.8  & 1.5  & gradient-based \\
SNAS + aggressive constraint + cutout~\citep{xie2018snas}  & 3.10 & -     & 2.3  & 1.5  & gradient-based \\
ProxylessNAS~\citep{cai2018proxylessnas} + cutout          & 2.08 & -     & 5.7  & 4.0  & gradient-based \\
\noalign{\smallskip}\hline\noalign{\smallskip}
P-DARTS (searched on CIFAR10) + cutout                     & 2.50 & 17.20 & 3.4  & 0.3 & gradient-based \\
P-DARTS (searched on CIFAR100) + cutout                    & 2.62 & 15.92 & 3.6  & 0.3 & gradient-based \\
P-DARTS (searched on CIFAR10, large) + cutout              & 2.25 & 15.27 & 10.5 & 0.3 & gradient-based \\
P-DARTS (searched on CIFAR100, large) + cutout             & 2.43 & 14.64 & 11.0 & 0.3 & gradient-based \\
\noalign{\smallskip}\hline
\end{tabular}
\end{center}
\label{tab_ev_cifar}
\end{table*}

\subsection{Relationship to Prior Work}

Though having a similar name, Progressive NAS or PNAS~\citep{liu2018progressive} was driven by a different motivation. PNAS explored the search space progressively by searching for operations node-by-node within each cell. Our approach shares a similar progressive search manner -- we perform the search at the cell level to enlarge the architecture depth, while PNAS did it at the operation level (within a cell) to reduce the number of architectures to evaluate.

There exist other efforts in alleviating the optimization gap between search and evaluation. For example, SNAS~\citep{xie2018snas} aimed at eliminating the bias between the search and evaluation objectives of differentiable NAS approaches by forcing the architecture weights on each edge to be one-hot. Our work is also able to get rid of the bias, which we achieve by enlarging the architecture depth during the search stage.

Another example of bridging the optimization gap is ProxylessNAS~\citep{cai2018proxylessnas}, which introduced a differentiable NAS scheme to directly learn architectures on the target task (and hardware) without a proxy dataset. It achieved high memory efficiency by applying binary masks to candidate operations and forcing only one path in the over-parameterized network to be activated and loaded into GPU. Different from it, our approach tackles the memory overhead by search space approximation. Besides, ProxylessNAS searched for global topology instead of cell topology, which requires strong priors on the target task as well as the search space, while P-DARTS does not need such priors. Our approach is much faster than ProxylessNAS ($0.3$ GPU-days vs. $4.0$ GPU-days on CIFAR10 and $2.0$ GPU-days vs. $8.3$ GPU-days on ImageNet).

Last but not least, we believe that the phenomenon that the \textit{skip-connect} operation emerges may be caused by the mathematical mechanism that DARTS was built upon. Some recent work~\citep{bi2019stabilizing} pointed out issues in optimization, and we look forward to exploring the relationship between these issues and the optimization gap.

\section{Experiments}\label{exp}

\subsection{The CIFAR10 and CIFAR100 Datasets}

Following standard vision setting, we search and evaluate architectures on the CIFAR10~\citep{krizhevsky2009learning} dataset. To further demonstrate the capability of our proposed method, we also execute architecture search on CIFAR100. 

Each of CIFAR10 and CIFAR100 has $50$K/$10$K training/testing images with a fixed spatial resolution of $32\times32$, which are distributed over $10$/$100$ classes.
In the architecture search scenario, the training set is randomly split into two equal subsets,
one for learning network parameters ({\em e.g.}, convolutional weights) and the other for tuning the architectural parameters ({\em i.e.}, operation weights).
In the evaluation scenario, standard training/testing split is applied.

\subsubsection{Architecture Search}\label{sr}

\begin{figure*}[t]
\centering
\begin{minipage}{0.32\textwidth}
\subfloat[Stage $1$, CIFAR10 Test Err. $2.90\%$]{\includegraphics[width=1.0\linewidth]{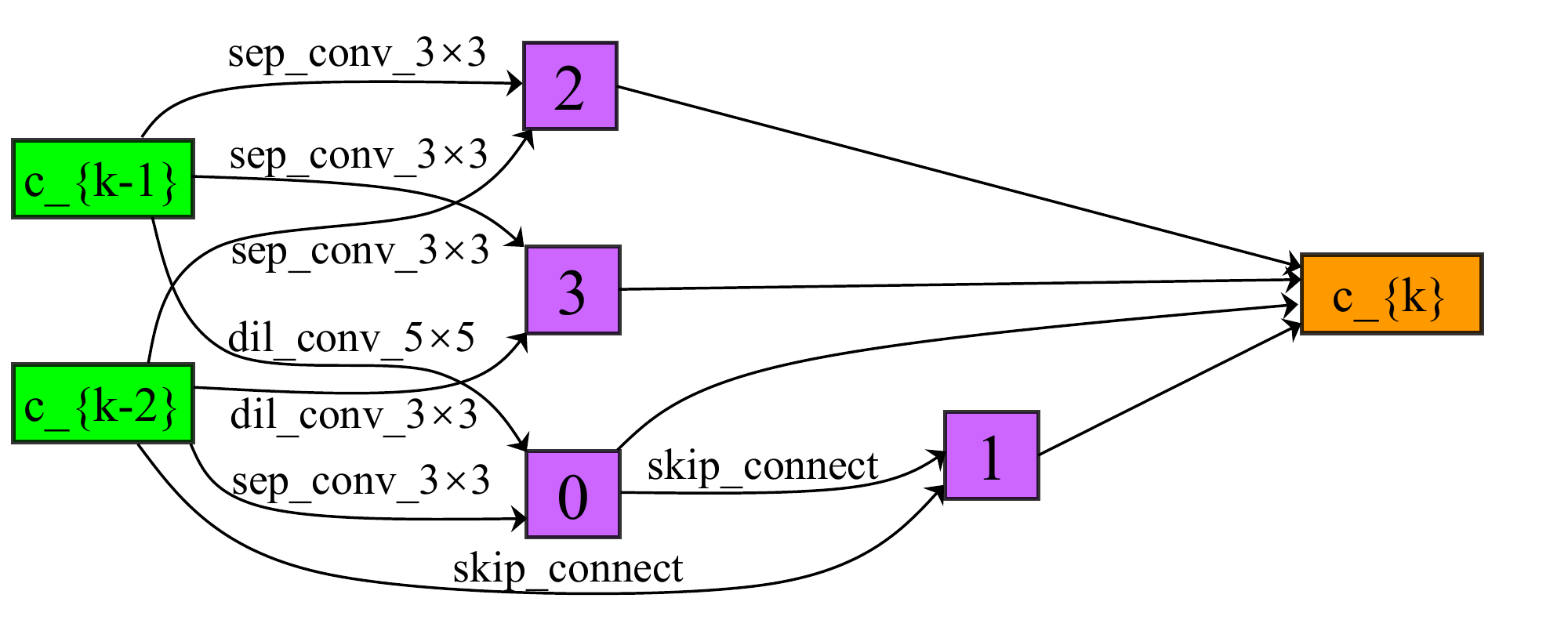}}\label{ncells_s1}\\
\subfloat[Stage $2$, CIFAR10 Test Err. $2.82\%$]{\includegraphics[width=1.0\linewidth]{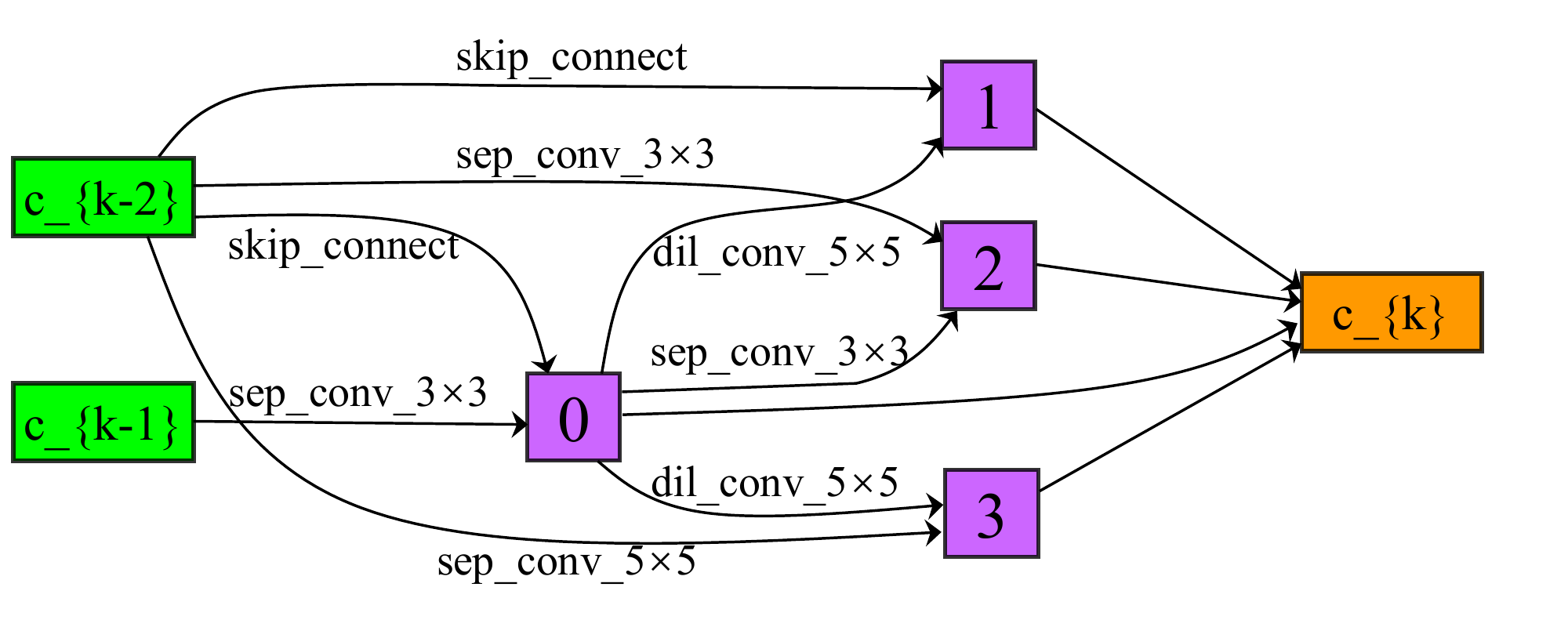}}\label{ncells_s2}
\end{minipage}
\begin{minipage}{0.32\textwidth}
\subfloat[Stage $3$, CIFAR10 Test Err. $2.58\%$]{\includegraphics[width=1.0\linewidth]{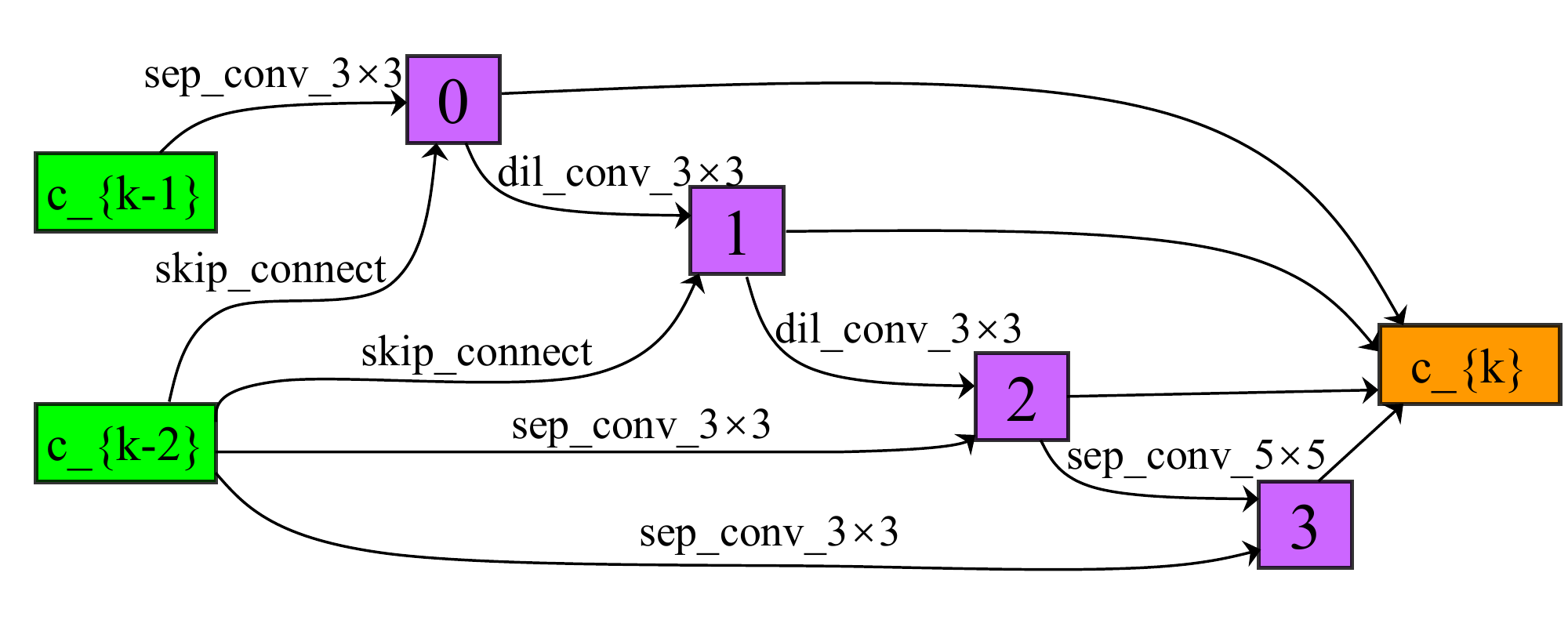}}\label{ncells_s3}\\
\subfloat[DARTS\_V2, CIFAR10 Test Err. $2.83\%$]{\includegraphics[width=1.0\linewidth]{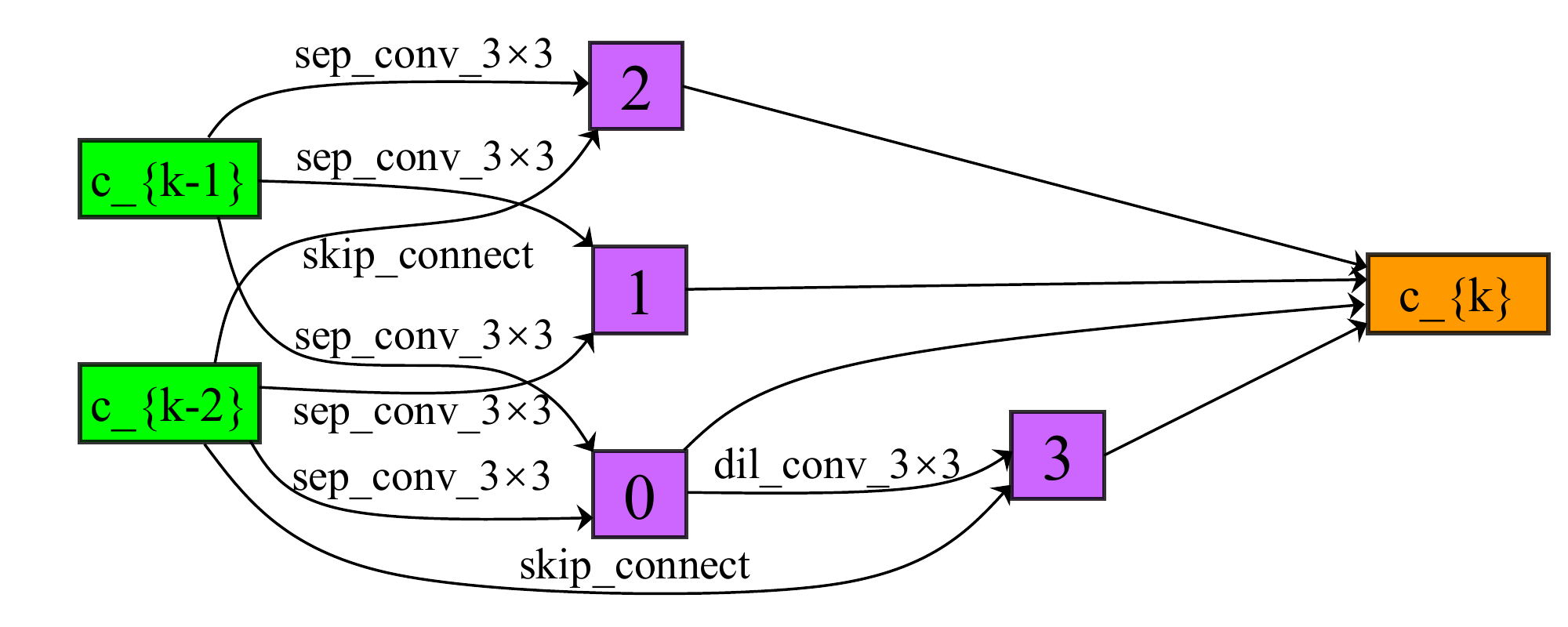}}\label{ncells_dv2}
\end{minipage}
\begin{minipage}{0.35\textwidth}
\subfloat[Statistics on connection levels]{\includegraphics[width=0.98\linewidth]{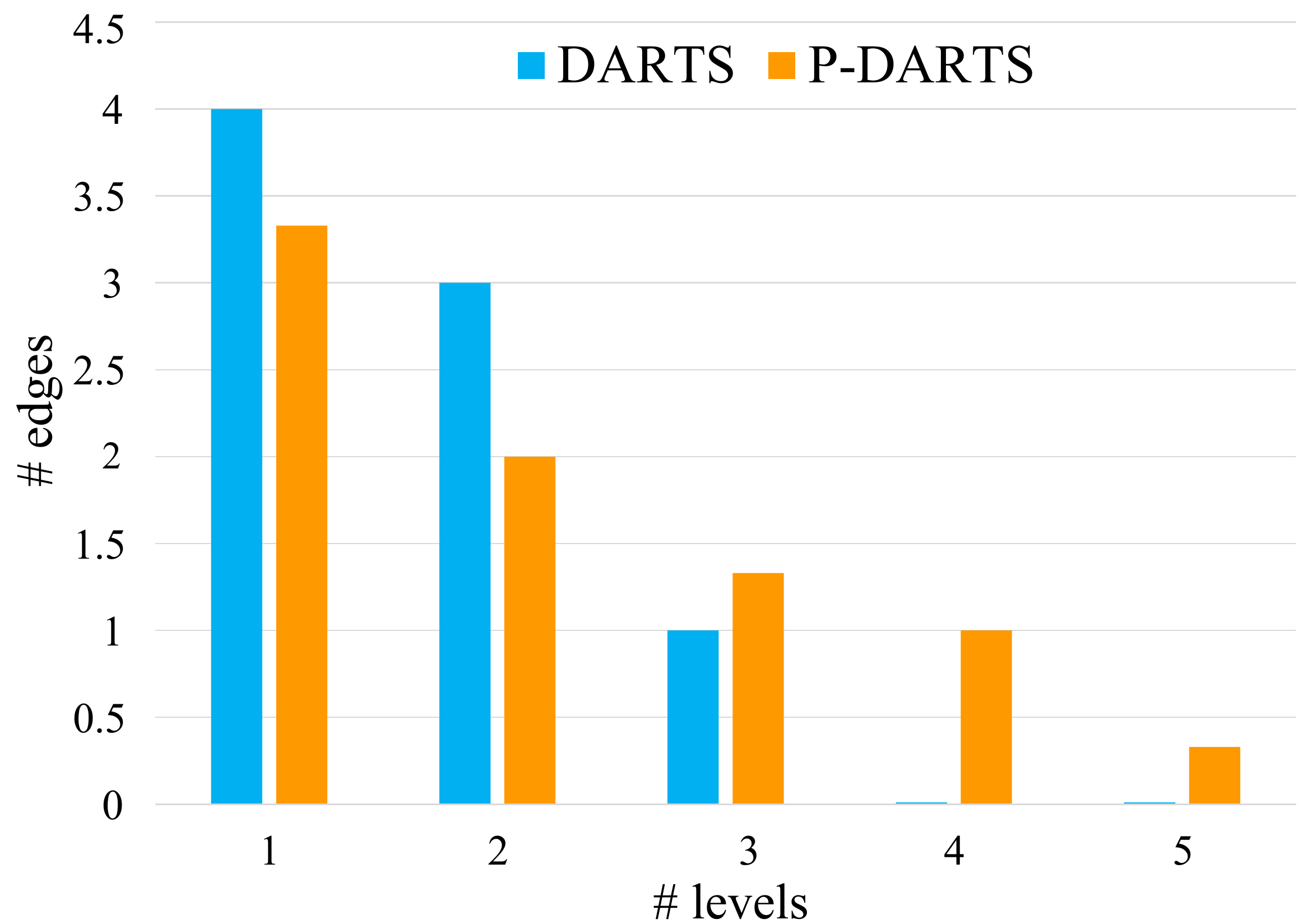}}\label{edge_level}\\
\end{minipage}
\caption{Left and middle: normal cells discovered by different search stages of P-DARTS and the second-order DARTS (DARTS\_V2). The depths of search networks are $5$, $11$ and $17$ cells for stage $1$, $2$ and $3$ of P-DARTS and $8$ for DARTS\_V2. When the depth of the search network increases, more deep connections are preserved. Note that the operation on edge $\mathrm{E}_{(0, 1)}$ of stage $1$ is a parameter-free \textit{skip\_connect}, thus it is strictly not a deep connection. Right: number of edges with different connection levels in the discovered architectures of DARTS and P-DARTS. More deep connections with higher connection levels are preserved in architectures discovered by P-DARTS, while only one exists in the architecture searched by DARTS. }
\label{ncells}
\end{figure*}

The whole search process is split into $3$ stages. 
The search space and network configuration are identical to DARTS at the initial stage (stage $1$) except that only $5$ cells are stacked in the search network for acceleration (we tried the original setting with $8$ cells and obtained similar results). 
The number of stacked cells increases from $5$ to $11$ for the intermediate stage (stage $2$) and $17$ for the final stage (stage $3$). The numbers of operations preserved on each edge of the super-network are $8$, $5$, and $3$ for stage $1$, $2$, and $3$, respectively. 

The Dropout probability on \textit{skip-connect} is decayed exponentially and the initial values for the reported results are set to be $0.0$, $0.4$, $0.7$ on CIFAR10 for stage $1$, $2$ and $3$, respectively, and $0.1$, $0.2$, $0.3$ for CIFAR100.
For a proper tradeoff between classification accuracy and computational overhead,
the final discovered cells are restricted to keep at most $2$ \textit{skip-connects}, which guarantees a fair comparison with DARTS and other state-of-the-art approaches.
For each stage, the super-network is trained for $25$ epochs with a batch size of $96$,
where only network parameters are tuned in the first $10$ epochs
while both network and architectural parameters are jointly optimized in the rest $15$ epochs. 
An Adam optimizer with learning rate $\eta = 0.0006$, weight decay $0.001$ and momentum $\beta=(0.5, 0.999)$ is adopted for architectural parameters. 
The limitation of GPU memory is the main concern when we determine hyper-parameters related to GPU memory size, {\em e.g.}, the batch size. 
The first-order optimization scheme of DARTS is leveraged to learn the architectural parameters in consideration of acceleration, thus the architectural parameters and network parameters are optimized in an alternative manner.  

The architecture search process on CIFAR10 and CIFAR100 is performed on a single Nvidia Tesla P100, which takes around $7$ hours, resulting in a search cost of $0.3$ GPU-days. When we change the GPU device to an Nvidia Tesla V100 ($16$GB), the search cost is reduced to $0.2$ GPU-days (around $4.5$ hours). 

Architectures discovered by P-DARTS on CIFAR10 tend to preserve more deep connections than the one discovered by DARTS, as shown in Figure~\ref{ncells}(c) and Figure~\ref{ncells}(d). 
Moreover, the deep connections in the architecture discovered by P-DARTS are deeper than that in DARTS, which means that the longest path in the cell cascades more levels in depth. In other words, there are more serial layers in the cell instead of parallel ones, making the target network further deeper and achieving better classification performance. 

Notably, our method also allows architecture search on CIFAR100 while prior approaches mostly failed.
The evaluation results in Table~\ref{tab_ev_cifar} show that the discovered architecture on CIFAR100 outperforms those architectures transferred from CIFAR10.
We tried to perform architecture search on CIFAR100 with DARTS using the code released by the authors
but get architectures full of \textit{skip-connects}, which results in much worse classification performance.

\subsubsection{Architecture Evaluation}

Following the convention\citep{liu2018darts}, an evaluation network stacked with $20$ cells and $36$ initial channels is trained from scratch for $600$ epochs with a batch size of $128$.
Additional regularization methods are also applied including Cutout regularization~\citep{devries2017improved, zhong2017random} of length $16$, drop-path~\citep{larsson2016fractalnet} of probability $0.3$ and auxiliary towers~\citep{szegedy2015going} of weight $0.4$.
A standard SGD optimizer with a momentum of $0.9$, a weight decay of $0.0003$ for CIFAR10 and $0.0005$ for CIFAR100 is adopted to optimize the network parameters.
The cosine annealing scheme is applied to decay the learning rate from $0.025$ to $0$. 
To explore the potential of the searched architectures, we further increase the number of initial channels from $36$ to $64$, which is denoted as the large setting.  

Evaluation results and comparison with state-of-the-art approaches are summarized in Table~\ref{tab_ev_cifar}.
As demonstrated in Table~\ref{tab_ev_cifar}, P-DARTS achieves a $2.50\%$ test error on CIFAR10 with a search cost of only $0.3$ GPU-days. 
To obtain a similar performance, AmoebaNet~\citep{real2018regularized} spent thousands of GPU-hours, which is four orders of magnitude more computational resources.
Our P-DARTS also outperforms DARTS and SNAS by a large margin with comparable parameter count. 
Notably, architectures discovered by P-DARTS outperform ENAS, the previously most efficient approach, in both classification performance and search cost, with fewer parameters.

The architectures discovered both DARTS and P-DARTS on CIFAR10 are transferred to CIFAR100 to test the transferability between similar datasets.
Obvious superiority of P-DARTS is observed in terms of classification accuracy. 
As mentioned previously, P-DARTS is able to support architecture search on other proxy datasets such as CIFAR100. 
For a fair comparison, we tried to perform architecture search on CIFAR100 with the publicly available code of DARTS but resulted in architectures full of \textit{skip-connect} operations. 
The discovered architecture on CIFAR100 significantly outperforms those architectures transferred from CIFAR10. 
An interesting point is that the directly searched architectures perform better when evaluated on the search dataset than those transferred ones for both CIFAR10 and CIFAR100.
Such a phenomenon provides a proof of the existence of dataset bias in NAS.

\subsection{Diagnostic Experiments} \label{exp_diag}

Before continuing to ImageNet search and in-the-wild evaluation experiments, we conduct diagnostic studies to better understand the properties of P-DARTS. 

\subsubsection{Comparison on the Depth of Search Networks}\label{comp_depth}

Since the search process of P-DARTS is divided into multiple stages, we perform a case study to extract architectures from each search stage with the same rule to validate the usefulness of bridging the depth gap.
Architectures from each stage are evaluated to demonstrate their capability for image classification.
The topology of discovered architectures (only normal cells are shown) and their corresponding classification performance are summarized in Figure~\ref{ncells}.
To show the difference in the topology of cells searched with different depth, we add the architecture discovered by second-order DARTS (DARTS\_V2, $8$ cells in the search network) for comparison.

The lowest test error is achieved by the architecture obtained from the final search stage (stage $3$), which validates the effectiveness of shrinking the depth gap.
From Figure~\ref{ncells} we can observe that these discovered architectures share some common edges,
for example \textit{sep\_conv\_$3\times3$} at edge $\mathrm{E}_{(c_{k-2}, 2)}$ for all stage of P-DARTS and at edge $\mathrm{E}_{(c_{k-1}, 0)}$ for stage $2$, $3$ of P-DARTS and DARTS$\_$V2. These common edges serve as a solid proof that operations with high importance are preserved by the search space approximation scheme. 
Differences also exist between these discovered architectures, which we believe is the key factor that affects the capability of these architectures.
Architectures generated by shallow search networks tend to keep shallow connections, while with deeper search networks, the discovered architectures prefer to pick some preceding intermediate nodes as input, resulting in cells with deep connections. 
This is because it is harder to optimize a deep search network, so the algorithm has to explore more paths to find the optimum, which results in more complex and powerful architectures. 

Additionally, we perform quantitative analysis on the discovered architectures by P-DARTS of three individual runs and summarize the average levels of their connections in Figure~\ref{ncells}(e). For comparison, we also add the architecture found by the second-order DARTS into this analysis. While the preserved edges of DARTS are almost all shallow ($7$ over $8$ of level $1$ and level $2$), P-DARTS tends to keep more deep edges. 

\subsubsection{Effectiveness of Search Space Approximation}

The search process takes $\sim7$ hours on a single Nvidia Tesla P100 GPU with $16$GB memory to produce the final architectures.
We monitor the GPU memory usage of the architecture search process for $3$ individual runs and collect the peak value to verify the effectiveness of the search space approximation scheme, which is shown in Table~\ref{mem_usage}. The memory usage is stably under the limit of the adopted GPU, and out of memory error barely occurs, showing the validity of the search space approximation scheme on memory efficiency. 

We perform experiments to demonstrate the effectiveness of the search space approximation scheme on improving classification accuracy.
Only the final stage of the search process is executed on two different search spaces with identical settings.
The first search space is progressively approximated from previous search stages, and the other is randomly sampled from the full search space.
To eliminate the influence of randomness, we repeat the whole process for the randomly sampled one for $3$ times with different seeds and pick the best one. 
The lowest test error for the randomly sampled search space is $3.43\%$, which is much higher than $2.58\%$, the one obtained with the approximated search space.
Moreover, we performed an additional experiment with a fixed depth ($8$ cells) and shrunk sets of operations ($8\to5\to3$, as used in the paper), which results in a test error of $2.70\%$, significantly lower than the $3.00\%$ test error obtained by the first-order DARTS. 
Such results reveal the necessity of the search space approximation scheme.

\begin{table}[!t]
\caption{Peak GPU memory usage at different stages during three individual runs. The memory limit is $16$GB.}
\begin{center}
\begin{tabular}{lccc}
\hline\noalign{\smallskip}
\multirow{2}{*}{\textbf{Run No.}} & \multicolumn{3}{c}{\textbf{Mem. Usage (GB)}} \\
\cmidrule(lr){2-4}
             & \textbf{Stage $1$}  & \textbf{Stage $2$} & \textbf{Stage $3$} \\
\noalign{\smallskip}\hline\noalign{\smallskip}
1  & 9.8 & 14.0 & 14.2 \\
2  & 9.8 & 14.4 & 14.5 \\
3  & 9.8 & 14.2 & 14.3 \\
\noalign{\smallskip}\hline
\end{tabular}
\end{center}
\label{mem_usage}
\end{table}

\begin{table*} [t]
\caption{Comparison with state-of-the-art architectures on ImageNet (mobile setting). $^\dagger$: the top-1 test error is $25.4\%$ when the learning rate decay schedule is cosine annealing.}
\begin{center}
\begin{tabular}{lcccccc}
\hline\noalign{\smallskip}
\textbf{\multirow{2}{*}{Architecture}} & \multicolumn{2}{c}{\textbf{Test Err. (\%)}} & \textbf{Params} & $\times+$ & \textbf{Search Cost} & \textbf{\multirow{2}{*}{Search Method}} \\
\cmidrule(lr){2-3}
&                            \textbf{top-1} & \textbf{top-5} & \textbf{(M)} & \textbf{(M)} & \textbf{(GPU-days)} &\\
\noalign{\smallskip}\hline\noalign{\smallskip}
Inception V1~\citep{szegedy2015going}          & 30.2 & 10.1 & 6.6 & 1\rm{,}448 & -    & manual \\
MobileNet~\citep{howard2017mobilenets}         & 29.4 & 10.5 & 4.2 & 569  & -    & manual \\
ShuffleNet V1 2$\times$~\citep{zhang2018shufflenet} & 26.4 & 10.2 & $\sim$5  & 524  & -    & manual \\
ShuffleNet V2 2$\times$~\citep{ma2018shufflenet}    & 25.1 & - & $\sim$5  & 591  & -    & manual \\
\noalign{\smallskip}\hline\noalign{\smallskip}
NASNet-A~\citep{zoph2018learning}              & 26.0 & 8.4  & 5.3 & 564  & 1\rm{,}800 & RL \\
NASNet-B~\citep{zoph2018learning}              & 27.2 & 8.7  & 5.3 & 488  & 1\rm{,}800 & RL \\
NASNet-C~\citep{zoph2018learning}              & 27.5 & 9.0  & 4.9 & 558  & 1\rm{,}800 & RL \\
AmoebaNet-A~\citep{real2018regularized}        & 25.5 & 8.0  & 5.1 & 555  & 3\rm{,}150 & evolution \\
AmoebaNet-B~\citep{real2018regularized}        & 26.0 & 8.5  & 5.3 & 555  & 3\rm{,}150 & evolution \\
AmoebaNet-C~\citep{real2018regularized}        & 24.3 & 7.6  & 6.4 & 570  & 3\rm{,}150 & evolution \\
PNAS~\citep{liu2018progressive}                & 25.8 & 8.1  & 5.1 & 588  & 225  & SMBO \\
MnasNet-92~\citep{tan2019mnasnet}              & 25.2 & 8.0  & 4.4 & 388  & -    & RL \\
\noalign{\smallskip}\hline\noalign{\smallskip}
DARTS (second order)~\citep{liu2018darts}      & 26.7$^\dagger$ & 8.7  & 4.7 & 574  & 4.0    & gradient-based \\
SNAS (mild constraint)~\citep{xie2018snas}     & 27.3 & 9.2  & 4.3 & 522  & 1.5  & gradient-based \\
PC-DARTS (searched on CIFAR10)~\citep{xu2019pc} & 25.1 & 7.8 & 5.3 & 586 & 0.1 & gradient-based \\
PC-DARTS (searched on ImageNet)~\citep{xu2019pc} & 24.2 & 7.3 & 5.3 & 597 & 3.8 & gradient-based \\
ProxylessNAS (GPU)~\citep{cai2018proxylessnas}       & 24.9 & 7.5  & 7.1 & 465  & 8.3  & gradient-based \\
\noalign{\smallskip}\hline\noalign{\smallskip}
P-DARTS (searched on CIFAR10)                 & 24.4 & 7.4  & 4.9 & 557  & 0.3  & gradient-based \\
P-DARTS (searched on ImageNet)                & 24.1 & 7.3  & 5.4 & 597  & 2.0  & gradient-based \\
\noalign{\smallskip}\hline
\end{tabular}
\end{center}
\label{ev_imagenet}
\end{table*} 

\subsubsection{Effectiveness of Search Space Regularization}

We perform experiments to validate the effectiveness of search space regularization, \textit{i.e.}, operation-level Dropout, and architecture refinement. 

\noindent\textbf{Effectiveness of operation-level Dropout}. 
Firstly, experiments are conducted to test the influence of the operation-level Dropout scheme.
Two sets of initial Dropout rates are adopted, \textit{i.e.}, $0.0$, $0.0$, $0.0$ (without Dropout) and $0.0$, $0.3$, $0.6$ (with Dropout) for stage $1$, $2$ and $3$, respectively.
To eliminate the potential influence of the number of \textit{skip-connects}, the comparison is made across multiple values of $M$.

Test errors for architectures discovered without Dropout are $2.93\%$, $3.28\%$ and $3.51\%$ for $M=2$, $3$ and $4$, respectively.
When operation-level Dropout is applied, the corresponding test errors are $2.69\%$, $2.84\%$ and $2.97\%$,
significantly outperforming those without Dropout.
According to the experimental results, all $8$ preserved operations in the normal cell of the architecture discovered without Dropout are \textit{skip-connects} before architecture refinement,
while the number is $4$ for the architecture discovered with Dropout.
The diminishing on the number of \textit{skip-connect} operations verifies the effectiveness of search space regularization on stabilizing the search process.

\noindent\textbf{Effectiveness of architecture refinement}. 
During experiments, we observe strong coincidence between the classification accuracy of architectures and the number of \textit{skip-connect} operations in them.
We perform a quantitative experiment to analyze it.
Architecture refinement is applied to the same search process to produce multiple architectures where the number of preserved \textit{skip-connect} operations in the normal cell varies from $0$ to $4$.

The test errors are positively correlated to the number of \textit{skip-connects} except for $M=0$, {\em i.e}, $2.78\%$, $2.68\%$, $2.69\%$, $2.84\%$ and $2.97\%$ for $M=0$ to $4$, while the parameters count is inversely proportional to the \textit{skip-connect} count, \textit{i.e.}, $4.1$M, $3.7$M, $3.3$M, $3.0$M and $2.7$M, respectively.
The reason lies in that, with a fixed number of operations in a cell,
the eliminated parameter-free \textit{skip-connects} are replaced by operations with trainable parameters, {\em e.g.}, \textit{convolution},
resulting in more complex and powerful architectures.

The above observation inspired us to propose the second search space regularization scheme, architecture refinement, whose capability is validated by the following experiments. 
We run another $3$ architecture search experiments, all with initial Dropout rates of $0.0$, $0.3$, and $0.6$ for stage $1$, $2$, and $3$, respectively.
Before architecture refinement, the test error is $2.79\pm0.16\%$ and the discovered architectures are with $2$, $3$ and $4$ \textit{skip-connects} in normal cells. 
After architecture refinement, all three searched architectures are with $2$ \textit{skip-connects} in normal cells, resulting in a diminished test error of $2.65\pm0.05\%$. 
The reduction of the average test error and standard deviation reveals the improvement of the stability for the search process. 

\noindent\textbf{Applying search space regularization to DARTS}.
We apply our proposed search space regularization scheme to the original first-order DARTS, and the test error on CIFAR10 is reduced to $2.78\%$, significantly lower than the original $3.00\%$ but still considerably higher than P-DARTS ($2.50\%$). This reveals that the proposed regularization scheme is also effective in searching for relatively shallower architectures, yet another source of improvement comes from increasing search depth. The positive results indicate that the proposed search space regularization can also be plugged into other DARTS-based approaches.  

\subsubsection{Discussion: Other Optimization Gaps}

Apart from depth gap that we have addressed in this paper, other aspects of the optimization gap can also affect the search process of super-network-based NAS approaches. Here, we briefly discuss two aspects of them.

\noindent\textbf{The width gap.} One of the straightforward option comes from the width of the network. Note that during the search stage on CIFAR, the base channel number is set to be $16$, while that is enlarged as $48$ when the searched architecture is transferred to ImageNet (see the experimental settings in the following section). This also claims a significant optimization gap.

Therefore, it is natural to progressively increase the network width during the search stage, just like what we have done for the network depth. However, we find that the performance gain brought by this strategy is limited. Digging into the searched architecture, we find that when an increased network width is used, the search algorithm tends to find architectures with small ($3\times3$) convolutional kernels, while the original version tends to find architectures with a considerable portion of big ($5\times5$) kernels. Consequently, the comparison between these two options is not fair on CIFAR10, as the original (not progressively widened) version often has a larger number of parameters. This also delivers an important message: the value of shrinking the optimization gap will be enlightened in a relatively ``fair''~\citep{chu2019fairnas} search environment.



\noindent\textbf{The gap brought by other hyper-parameters.} In the search setting of DARTS, all the affine parameters of batch normalization are discarded because the architectural parameters are dynamically learned across the whole search process, and the affine parameters will rescale the output of each operation according to incorrect statistics. On the contrary, the affine option of batch normalization is switched on to recover the data distribution during the evaluation scenario, which forms another aspect of the optimization gap. However, this gap is hard to address because a bunch of additional issues may arise if we switch it on. 

Furthermore, the data augmentation gap, including the inconsistent settings Cutout is another inconsistency between search and evaluation. There also may exist other aspects of the optimization gap, {\em e.g.}, Dropout, auxiliary loss tower, {\em etc.}. ~\citep{bi2019stabilizing} briefly discussed some aspects of the above-mentioned ones, while the influence of these options was not clearly stated. In fact, a different setting on these aspects may cause other additional problems to disrupt qualitative and quantitative analysis on them. Additionally, the fluctuation on small scale datasets like CIFAR10 may also cause dramatic impacts on the analysis, while the computational burden obstructs the analysis on large-scale datasets.

\subsection{The ImageNet Dataset}

We also search architectures directly on ImageNet to validate the applicability of our search algorithm to large-scale datasets. Experiments are performed on ILSVRC2012~\citep{russakovsky2015imagenet}, a subset of ImageNet~\citep{deng2009imagenet} which contains $1\rm{,}000$ object categories and $1.28$M training and $50$K validation images. 
Following the conventions~\citep{zoph2018learning, liu2018darts, wu2019fbnet}, we randomly sample a $100$-class subset of the training images for architecture search. Similar to CIFAR10, all images and standard dataset partition are adopted during architecture evaluation. 

\subsubsection{Architecture Search}

We use a similar configuration to the one used on CIFAR10 except for some minor changes. 
We set the numbers of cells to be $5$, $8$ and $11$ and adjust the dropout rate to $0.0$, $0.3$, $0.6$. 
Meanwhile, we increase the number of initial channels from $16$ to $28$, and $40$ for stage $1$, $2$, and $3$, respectively. 

Architecture search on ImageNet is executed with $8$ Nvidia Tesla V100, which
takes around $6$ hours, thus a search cost of $2$ GPU-days. The search cost of P-DARTS on ImageNet is even smaller than PC-DARTS\citep{xu2019pc}, a memory-efficient differentiable approach proposed recently, which demonstrates the efficiency of our proposed search space approximation scheme. 

During the search process, the ``over-fitting" phenomenon is largely alleviated and the number of \textit{skip-connect} operation is well controlled. This comes from two aspects. On the one hand, gradients assigned to \textit{skip-connects} is successfully suppressed by the first search space regularization method, {\em i.e.}, adding operation level dropout on \textit{skip-connect} operations. On the other hand, the variety and complexity of ImageNet make it more difficult to fit with those parameter-free operations than CIFAR10 and CIFAR100, forcing the network to train those operations with learnable parameters. 
Moreover, the discovered architecture is also with plenty of deep connections, even deeper than those in the one searched on CIFAR10. Such a character guarantees a favorable classification performance. We have also attempted to search architectures without progressively increasing the network width, but the discovered architectures resulted in worse classification performance, which demonstrates the usefulness of the progressive width scheme.

\begin{table*}[ht]
\caption{Detection results on the MS-COCO dataset (test-dev). $^\dagger$ denotes the results in this line are from~\citep{duan2019centernet}. The displayed FLOPs only includes the computations in the network backbone.}
\centering
\begin{tabular}{lccccccccc}	
\hline\noalign{\smallskip}
\textbf{Network}       &\textbf{Input Size}&\textbf{Backbone} &\textbf{$\times+$}&\textbf{$\mathrm{AP}$}&\textbf{$\mathrm{AP}_{50}$}  &\textbf{$\mathrm{AP}_{75}$}&\textbf{$\mathrm{AP}_\mathrm{S}$}&\textbf{$\mathrm{AP}_\mathrm{M}$} &\textbf{$\mathrm{AP}_\mathrm{L}$}\\
\noalign{\smallskip}\hline\noalign{\smallskip}
SSD300~\citep{liu2016ssd} 	&300$\times$300&VGG-16&31.4B&23.2&41.2&23.4&5.3 &23.2& 39.6\\
SSD512~\citep{liu2016ssd} 	&512$\times$512&VGG-16&80.4B&26.8&46.5&27.8&9.0 &28.9& 41.9\\
SSD513~\citep{liu2016ssd}$^\dagger$ &513$\times$513&ResNet-101& 43.4B &31.2&50.4&33.3&10.2&34.5&49.8\\
\noalign{\smallskip}\hline\noalign{\smallskip}
SSDLiteV1~\citep{howard2017mobilenets}   &320$\times$320&MobileNetV1&1.2B&22.2&-&-&-&-&-\\
SSDLiteV2~\citep{sandler2018mobilenetv2}&320$\times$320&MobileNetV2&0.7B&22.1&-&-&-&-&-\\
SSDLiteV3~\citep{tan2019mnasnet} &320$\times$320&MnasNet-A1&0.6B &23.0 &-& -&3.6 &20.5 &43.2\\
\noalign{\smallskip}\hline\noalign{\smallskip}
SSD320~\citep{liu2016ssd}   &320$\times$320&  DARTS           &1.1B &27.3&45.0&28.3&7.6&30.2&46.0\\
SSD320~\citep{liu2016ssd}   &320$\times$320&P-DARTS (CIFAR10) &1.1B &28.9&46.8&30.2&7.3&32.2&48.2\\
SSD320~\citep{liu2016ssd}   &320$\times$320&P-DARTS (ImageNet)&1.2B  &29.9&47.8&31.5&9.0&33.2&50.0\\
SSD512~\citep{liu2016ssd}   &512$\times$512& DARTS            &2.9B &31.8&50.3&33.8&11.7&37.1&49.7\\
SSD512~\citep{liu2016ssd}   &512$\times$512&P-DARTS (CIFAR10) &2.9B &33.6&52.8&35.6&13.3&39.7&51.1\\
SSD512~\citep{liu2016ssd}   &512$\times$512&P-DARTS (ImageNet)&3.1B &34.1&52.9&36.3&14.3&40.0&52.1\\
\noalign{\smallskip}\hline	
\end{tabular}
\label{tab_od}
\end{table*}

\subsubsection{Architecture Evaluation}

The transferability to large-scale recognition datasets of architecture discovered on CIFAR10 is firstly tested on The ILSVRC2012. Concurrently, the capability of the architecture directly searched on ImageNet is also evaluated. 
We apply the mobile setting for the evaluation scenario where the input image size is $224\times224$, and the number of multi-add operations is restricted to be less than $600$M.
A network configuration identical to DARTS is adopted, \textit{i.e.}, an evaluation network of $14$ cells and $48$ initial channels.
We train each network from scratch for $250$ epochs with batch size $1\rm{,}024$ on $8$ Nvidia Tesla V100 GPUs,
which takes about $3$ days with our PyTorch~\citep{paszke2017automatic} implementation.
The network parameters are optimized using an SGD optimizer with an initial learning rate of $0.5$ (decayed linearly after each epoch), a momentum of $0.9$, and a weight decay of $3\times 10^{-5}$.
Additional enhancements, including label smoothing~\citep{szegedy2016rethinking} and auxiliary loss tower, are applied during training. 
Since large batch size and learning rate are adopted, we apply learning rate warmup~\citep{goyal2017accurate} for the first $5$ epochs.

Evaluation results and comparison with state-of-the-art approaches are summarized in Table~\ref{ev_imagenet}.
The architecture transferred from CIFAR10 outperforms DARTS, PC-DARTS and SNAS by a large margin in terms of classification performance,
which demonstrates superior transfer capability of the discovered architectures. 
Notably, architectures discovered by P-DARTS on CIFAR10 and CIFAR100 achieve lower test error than MnasNet~\citep{tan2019mnasnet} and ProxylessNAS~\citep{cai2018proxylessnas}, whose search space is carefully designed for ImageNet. 
The architecture directly searched on ImageNet achieves superior performance compared to those transferred ones and is comparable to the state-of-the-art directly-searched models in the DARTS-based search space. 

\subsection{Evaluation in the Wild}


To further test the transferability of the discovered architectures to scenarios in the wild, we embed our discovered architectures as backbones into two other vision tasks, \textit{i.e.}, object detection and person re-identification. On both tasks, we have observed superior performance compared to both baseline methods and the DARTS backbone, which reveals that the desirable characters obtained on image recognition by P-DARTS can be well transferred to scenarios in the wild. 

\subsubsection{Transferring to Object Detection}\label{transfer_od}

\begin{table*} [t]
\caption{Results of person re-identification on Market-1501, DukeMCMT-reID and MSMT17. The displayed FLOPs only includes the computations in the network backbone.}
\begin{center}
\begin{tabular}{lccccccccc}
\hline\noalign{\smallskip}
\textbf{\multirow{2}{*}{Backbone}} & \textbf{\multirow{2}{*}{\# Parts}} & \textbf{\multirow{2}{*}{Feat. Dim}} & \textbf{$\times+$} &  \multicolumn{2}{c}{\textbf{Market-1501}} & \multicolumn{2}{c}{\textbf{DukeMCMT-reID}} & \multicolumn{2}{c}{\textbf{MSMT17}} \\
\cmidrule(lr){5-6} \cmidrule(lr){7-8} \cmidrule(lr){9-10} 
 & & & \textbf{(M)} &\textbf{Rank-$1$} & \textbf{mAP} & \textbf{Rank-$1$} & \textbf{mAP} & \textbf{Rank-$1$} & \textbf{mAP}\\
\noalign{\smallskip}\hline\noalign{\smallskip}
ResNet-50           & 1 & 2,048 & 4120 & 87.86 & 72.8  & 71.99 & 57.21 & 48.33 & 24.62\\
DARTS               & 1 &  768  & 573  & 91.90 & 79.3  & 82.09 & 66.74 & 61.50 & 37.93\\
P-DARTS (CIFAR10)   & 1 &  768  & 556  & 92.99 & 81.37 & 83.75 & 68.71 & 68.98 & 41.99\\
P-DARTS (ImageNet)  & 1 &  768  & 596  & 92.01 & 78.41 & 83.55 & 67.84 & 66.70 & 39.64\\
\noalign{\smallskip}\hline\noalign{\smallskip}
ResNet-50          & 3 & 2,048 & 4120 & 92.81 & 80.34 & 84.61 & 71.51 & 71.64 & 46.23\\
DARTS              & 3 &  768  & 573  & 94.18 & 83.63 & 86.22 & 74.68 & 77.37 & 53.02\\
P-DARTS (CIFAR10)  & 3 &  768  & 556  & 94.59 & 84.78 & 87.25 & 75.53 & 79.52 & 55.98\\
P-DARTS (ImageNet) & 3 &  768  & 596  & 93.67 & 83.85 & 89.98 & 75.27 & 77.19 & 53.37\\
\noalign{\smallskip}\hline\noalign{\smallskip}
ResNet-50          & 6 & 2,048 & 4120 & 93.08 & 81.02 & 86.13 & 74.03 & 71.12 & 46.00\\
DARTS              & 6 &  768  & 573  & 93.40 & 83.23 & 86.35 & 74.22 & 77.14 & 53.84\\
P-DARTS (CIFAR10)  & 6 &  768  & 556  & 93.61 & 83.37 & 87.25 & 74.6  & 79.24 & 56.41\\
P-DARTS (ImageNet) & 6 &  768  & 596  & 92.99 & 82.98 & 86.71 & 73.96 & 76.75 & 53.62\\
\hline\noalign{\smallskip}
\end{tabular}
\end{center}
\label{tab_reid}
\end{table*}

Object detection is also a fundamental task in the vision community and also an important task of the scenario in the wild~\citep{Liu2019}. We plug the discovered architectures and corresponding weights pre-trained on ImageNet into Single-Shot Detectors (SSD)~\citep{liu2016ssd}, a popular light-weight object detection framework. The capability of our backbones is tested on the benchmark dataset MS-COCO~\citep{lin2014microsoft}, which contains $80$ object categories and more than $1.5$M object instances. We train the pipeline with the ``trainval$35$K'' set, \textit{i.e.}, a combination of the $80$k training and $35$k validation images. The performance is tested on the \textit{test-dev} set.

Results are summarized in Table~\ref{tab_od}. Equipped with the P-DARTS backbone searched on CIFAR10, the P-DARTS-SSD$320$ achieves an superior AP of $28.9\%$ with only $1.1$B FLOPs, which is $5.7\%$ higher in AP with $29\times$ fewer FLOPs than SSD$300$, and even $2.1$\% higher in AP with $73\times$ fewer FLOPS than the SSD$512$. With similar FLOPs, P-DARTS-SSD$320$ outperforms the DARTS-SSD$320$ by $1.6\%$ in AP. Compared to those light-weight backbones, {\em i.e.}, backbones belong to the MobileNet family, our P-DARTS-SSD$320$ enjoys a superior AP by a large margin, while with an acceptable amount of extra FLOPs than these light-weight backbones. With larger input image size, the P-DARTS-SSD$512$ surpasses the SSD$513$ by an AP of $2.4\%$, while the FLOPs count of the P-DARTS backbone is $14\times$ smaller than their ResNet-$101$ backbone. 
The results with the backbone searched by P-DARTS on ImageNet are further impressive, which achieves an AP of $29.9\%$ with the backbone searched on CIFAR10 for P-DARTS-SSD$320$, and $34.1\%$ for P-DARTS-SSD$512$. 
All the above results indicate that the discovered architectures by P-DARTS are well transferred to object detection and produce superior performance.

\subsubsection{Transferring to Person Re-Identification}

Person re-identification is an important practical vision task and has been attracting more attention from both academia and industry~\citep{wang2018person, li2018semi} because of its broad applications in surveillance and security. Apart from those task-specific modules, the backbone architecture is a critical factor for performance promotion. We replace the previous backbones with our P-DARTS architectures (searched on both CIFAR10 and ImageNet) and test the transferability on three benchmark datasets, \textit{i.e.}, \textit{Market-1501}~\citep{zheng2015scalable}, \textit{DukeMTMC-reID}~\citep{zheng2017unlabeled} and \textit{MSMT17}~\citep{wei2018person}. Experiments are executed with the pipeline of Part-based Convolutional Baseline (PCB)~\citep{sun2018beyond}, and all backbones are pre-trained on ImageNet. We set the numbers of parts to be $1$, $3$, and $6$ to make an exhaustive comparison. 

Results are summarized in Table~\ref{tab_reid}. The P-DARTS backbones outperform the ResNet-$50$ backbone by a large margin with fewer FLOPs and a smaller feature dimensionality. With a similar backbone size, P-DARTS (CIFAR10) surpasses DARTS on all three datasets with different part numbers, suggesting a superior transferability of our searched architecture. However, with the P-DARTS (ImageNet) backbone, the performance is only comparable to the DARTS backbone and worse than the P-DARTS (CIFAR10) backbone.

\begin{figure}[tb]
\subfloat[ImageNet]{
\begin{minipage}[t]{0.49\linewidth}
\centering
\includegraphics[width=1.5in]{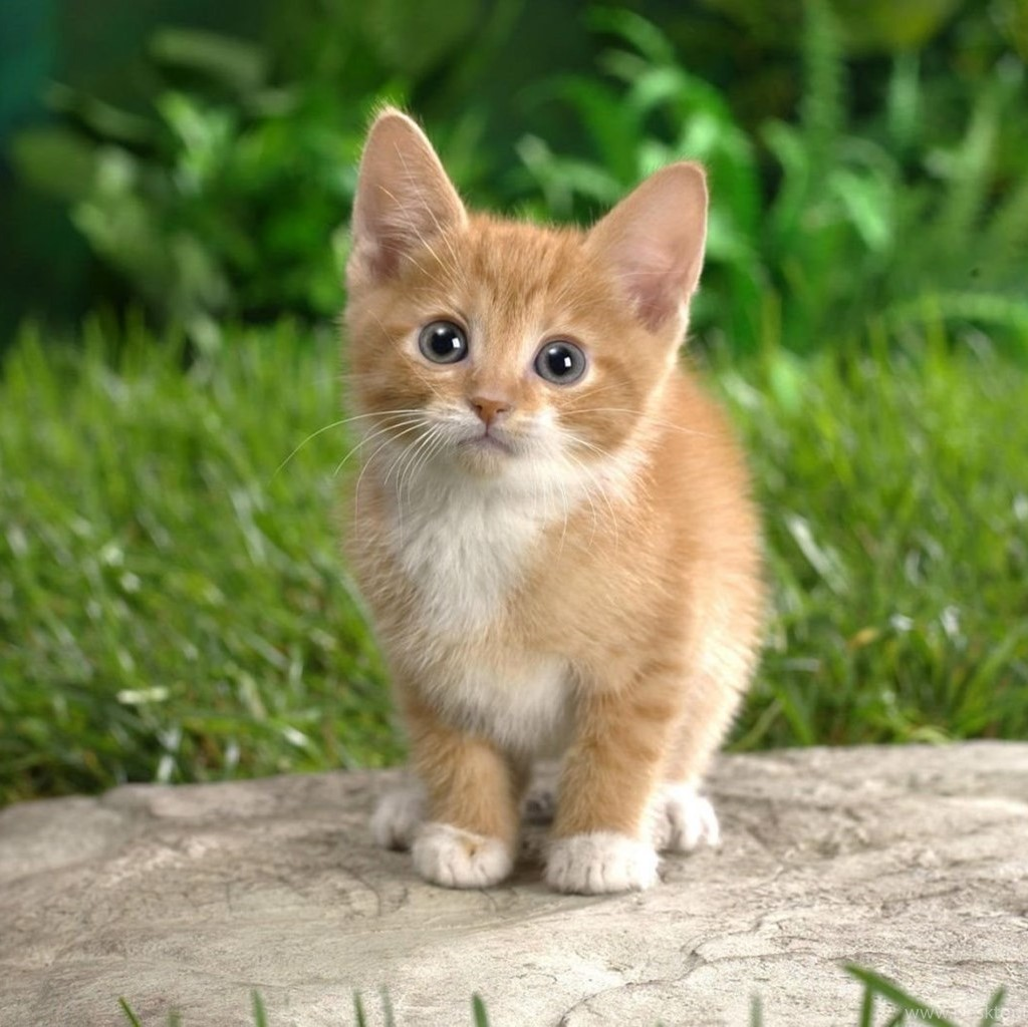}
\end{minipage}%
}%
\subfloat[MS-COCO]{
\begin{minipage}[t]{0.49\linewidth}
\centering
\includegraphics[width=1.5in]{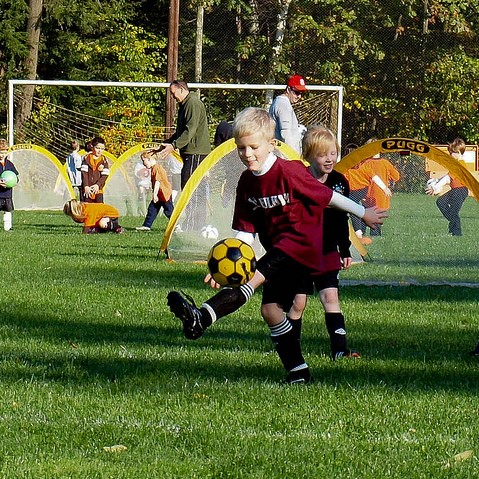}
\end{minipage}%
}%
\quad
\subfloat[CIFAR10]{
\begin{minipage}[t]{0.49\linewidth}
\centering
\includegraphics[width=1.5in]{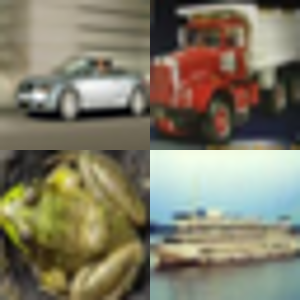}
\end{minipage}
}%
\subfloat[Market-1501]{
\begin{minipage}[t]{0.49\linewidth}
\centering
\includegraphics[width=1.5in]{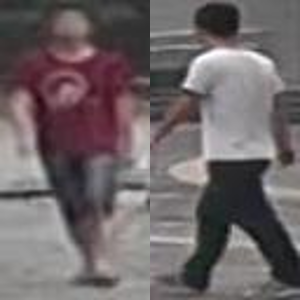}
\end{minipage}
}%
\caption{Samples from ImageNet, MS-COCO, CIFAR10 and Market-1501. Please zoom in to see clearer.}
\label{data_samples}
\end{figure}

It is worth noting that the preferences to CIFAR-searched and ImageNet-searched backbones are different between object detection and person re-identification tasks. This is due to the domain gap between the architecture search task and the target tasks. While the original images used in ImageNet classification and COCO object detection are similarly with high resolution and data distribution, images used in ReID are in worse condition, which is more similar to the situation in CIFAR10. We showcase in Figure~\ref{data_samples} samples from ImageNet, COCO, CIFAR10, and Market-1501, where the domain gap between them can be visually observed. 
A notable phenomenon is that with the ResNet-$50$ backbone, performance keeps rising when increasing the part number, while the best performance is reached to the peak when the part number is $3$ with both DARTS and P-DARTS backbones. This is arguably because of the larger feature dimensionality adopted in ResNet-$50$ backbone, which also implies the potential of further performance promotion on P-DARTS backbones with a larger feature dimensionality and part number. 

\section{Conclusions}\label{conclusion}

In this work, we propose a progressive version of differentiable architecture search to bridge the optimization gap between search and evaluation scenarios for NAS in the wild. The core idea, based on that optimization gap is caused by the difference between the policies of search and evaluation, is to gradually increase the depth of the super-network during the search process. To alleviate the issues of computational overhead and instability, we design two practical techniques to approximate and regularize the search process, respectively. Our approach reports superior performance in both proxy datasets (CIFAR and ImageNet) and target datasets (object detection and person re-identification added) with significantly reduced search overheads.

Our research puts forward the optimization gap in super-network-based NAS and highlights the significance of the consistency between search and evaluation scenarios. To solve it in terms of network depth and width, the P-DARTS algorithm paves a new way by approximating the search space. We expect that our initial work serves as a modest spur to induce more researchers to contribute their ideas to further alleviate the optimization gap and design effective and generalized NAS algorithms. 


\section*{Acknowledgement}
We thank Chen Wei, Jian Zhang, Kaiwen Duan, Longhui Wei, Tianyu Zhang, Yuhui Xu and Zhengsu Chen for their valuable suggestions.


%
%

\bibliographystyle{spbasic}      
\bibliography{egbib}   

%
%

\end{document}